\documentclass[lettersize,journal]{IEEEtran}
\usepackage{amsmath,amsfonts}
\usepackage{algorithm}
\usepackage{array}
\usepackage[caption=false,font=normalsize,labelfont=sf,textfont=sf]{subfig}
\usepackage{textcomp}
\usepackage{url}
\usepackage{verbatim}
\usepackage{graphicx}
\usepackage{cite}
\usepackage{algpseudocode}
\usepackage{newfloat}
\usepackage{listings}
\usepackage{multirow}
\usepackage{bibentry}
\usepackage{float}
\usepackage{dblfloatfix}
\usepackage{xcolor}
\usepackage{graphicx}
\usepackage{bbm}
\usepackage{mathtools}
\usepackage{enumitem}
\usepackage{amsmath}
\usepackage{booktabs}

\usepackage{CJKutf8}

\begin{document}

\title{Only Send What You Need: \\ Learning to Communicate Efficiently in \\ Federated Multilingual Machine Translation}


\newcommand{\yun}[1]{{\small\color{blue}{\bf\xspace#1 -yun}}}
\newcommand{\cgb}[1]{{\color{red}{cgb: \xspace#1}}}

\author{Yun-Wei Chu, Dong-Jun Han, Christopher G. Brinton

\thanks{Y.-W. Chu and C. Brinton are with the Elmore Family School of Electrical and Computer Engineering, Purdue University, IN, USA.email: chu198@purdue.edu, cgb@purdue.edu}
\thanks{D.-J. Han is with the Department of Computer Science and Engineering, Yonsei University, Korea. email: djh@yonsei.ac.kr}
\thanks{An abridged version \cite{Chu2024OnlySW} of this paper has been published in the Companion Proceedings of the ACM on Web Conference 2024.}

}

\markboth{IEEE/ACM Transactions on Audio, Speech, and Language Processing}%
{Shell \MakeLowercase{\textit{et al.}}: A Sample Article Using IEEEtran.cls for IEEE Journals}


\maketitle

\begin{abstract}
Federated learning (FL) is a promising distributed machine learning paradigm that enables multiple clients to collaboratively train a global model. 
In this paper, we focus on a practical federated multilingual learning setup where clients with their own language-specific data aim to collaboratively construct a high-quality neural machine translation (NMT) model. 
However, communication constraints in practical network systems present challenges for exchanging large-scale NMT engines between FL parties. 
We propose a meta-learning-based adaptive parameter selection methodology, \textbf{\texttt{MetaSend}}, that improves the communication efficiency of model transmissions from clients during FL-based multilingual NMT training.
Our approach learns a dynamic threshold for filtering parameters prior to transmission without compromising the NMT model quality, based on the tensor deviations of clients between different FL rounds.
Through experiments on two NMT datasets with different language distributions, we demonstrate that \texttt{MetaSend} obtains substantial improvements over baselines in translation quality in the presence of a limited communication budget.
\end{abstract}


\section{Introduction}
\label{sec:intro}

\begin{figure}[t]
    \vspace{-1.5mm}
    \centering
\setlength{\abovecaptionskip}{1mm}    \includegraphics[width=0.88\linewidth]{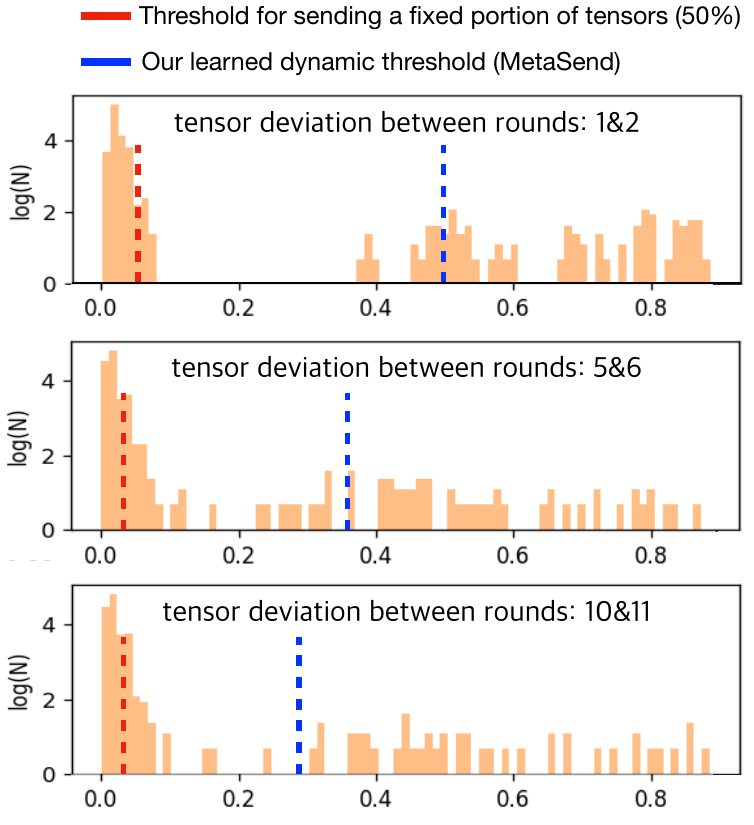}
    \caption{Sample histograms of the difference (absolute-value norms) between tensors of NMT engines computed for clients across consecutive communication rounds in FL training. The traditional method (red thresholds) fails to accurately capture the boundary between clusters during sending, while our \texttt{MetaSend} (blue thresholds) provides a dynamic threshold that adapts to the varying distribution across FL rounds.}
    \label{fig:clientdiff}
    \vspace{-4mm}
\end{figure}
\IEEEPARstart{F}{ederated} learning (FL) has recently emerged as a popular distributed machine learning paradigm as it
 enables collaborative model training among a set of clients via periodic aggregations of local models by a server~\cite{McMahan2017CommunicationEfficientLO, Konecn2016FederatedLS}.
The standard operation of FL is that each client sends its model parameters to the server, which periodically aggregates these parameters into a global model and synchronizes the global model to all clients in each communication round.
The FL property of keeping data local to clients has important privacy advantages that have made it attractive for many learning applications.

Natural language processing (NLP) is one domain standing to benefit from FL since user-generated text may contain sensitive information. Among the applications of FL in NLP, relatively few prior works have considered \textit{multilingual} NLP and the impact of different languages on FL~\cite{Liu2021FederatedLM} despite their practical significance. In recent years, neural machine translation (NMT)~\cite{Zhang2016AbstractiveCS,Zhang2017ACR,Wang2018SentenceSA} has shown substantial progress in this domain with the advent of large-scale language models such as BERT~\cite{devlin-etal-2019-bert}, GPT~\cite{Radford2019LanguageMA}, and their extensions.  NMT has a further natural alignment with FL given its setting of non-IID local data distributions~\cite{weller-etal-2022-pretrained}: each client (user) typically has a specific language direction they are interested in for translation, which their local dataset will be skewed towards, motivating them to collaborate with each other via FL to construct the general NMT model.

{However, multilingual NMT presents unique challenges for FL compared with more generic learning tasks. The presence of language heterogeneity, due to variations in vocabulary, grammar structures, and syntax across clients and languages, complicates model training \cite{weller-etal-2022-pretrained}. Dialects and regional variations add further complexity, as standard models may struggle to generalize across diverse linguistic patterns \cite{Liu2021FederatedLM}. Dataset imbalance also manifests in unique ways for NMT tasks, with certain languages being underrepresented across clients and having significantly fewer resources devoted to them compared to others. Together, these factors can lead to performance disparities for different clients and languages.}

Additionally, resource utilization is often a concern in deploying large-scale NMT models due to demands imposed on computational and memory resources~\cite{ Gupta2020CompressionOD}.
While FL will distribute the processing load, every client must exchange its model parameters with a central server during the FL communication phases. Communication efficiency is a known bottleneck in traditional FL applications~\cite{McMahan2017CommunicationEfficientLO} and becomes an even more critical challenge with large-scale NMT models.

\vspace{-2mm}
\subsection{Communication Efficiency in Federated NMT}
{In this paper, we are interested in addressing the aforementioned challenges simultaneously, i.e., optimizing multilingual NMT performance over an FL system with a limited communication budget.
While many existing works have studied communication-efficient FL for general learning tasks \cite{Diao2020HeteroFLCA, Sattler2019RobustAC}, federated multilingual NMT is nuanced due to the language heterogeneity, dialects, and imbalance characteristics mentioned above. The recent work \cite{passban-etal-2022-training} was the first to explore this, arguing that exchanging complete NMT engines in FL may be unnecessary due to how deviations in parameters/tensors tend to be distributed across consecutive training rounds.

To build intuition around this, we conduct a small NMT FL experiment using the well-known \texttt{FedAVG} algorithm~\cite{McMahan2017CommunicationEfficientLO}.
In Figure~\ref{fig:clientdiff}, we perform FL on the UN Corpus dataset (see Section~\ref{sec:exp_design} for details) distributed across three clients (each containing one language translation direction), and plot the differences in NMT model tensors between a few consecutive training rounds for one of the clients. 
These differences are computed and visualized tensor-by-tensor, showing deviations for each tensor.
We observe that most NMT model tensors have small deviations, while a small subset of tensors exhibit significant deviations. We find this to be the case across other clients, datasets, and distributions as well. This motivates parameter selection methods that send only specific tensors during each FL training round, with the selection tailored to this phenomenon of clustered deviations in federated NMT.

The method in \cite{passban-etal-2022-training} aims to account for this by focusing on transmitting either highly fluctuating or less active tensors during FL rounds to reduce communication load.
Their approach involves sending a fixed portion of model parameters -- namely, by sending either the top 50\% or the bottom 50\% of tensors based on their deviation, which is computed using the previous round's NMT engine.
However, it is important to note that sending a fixed portion of the parameters does not account for the fact that the deviation distribution will likely vary dynamically across rounds (as also observed in Figure~\ref{fig:clientdiff}).
As a result, this approach does not pay careful attention to the tradeoff between communication efficiency and NMT quality, potentially resulting in the transmission of either too many parameters, i.e., extra communication burden without any significant change in translation quality, or too few parameters, i.e., leading to an undesirable model that negatively impacts translation quality. Motivated by this, we pose the following research question: \textbf{\textit{How can we adaptively adjust the parameters transmitted across training rounds in federated NMT to reduce communication burden while not compromising the translation model quality?}}

To address this question, we focus on developing a \textit{dynamic} thresholding technique that can be applied for NMT model transmission during FL.
The central challenge involved is how to adaptively determine a threshold that selectively filters out parameters from transmission until we expect that the translation quality will start to be compromised.
Moreover, our approach aims to address the sensitivity of NMT quality and mitigate the risk of either excessive communication burden or inadequate parameter exchange.
To this end, we propose \textbf{\texttt{MetaSend}}, a model-agnostic meta-learning (MAML)-based methodology that generates a dynamic transmission threshold adapting to the varying tensor deviation distributions across training rounds. 
By crafting the MAML module adaptation according to the observed NMT quality across clients, our method accounts for the tradeoff between translation quality and communication efficiency in parameter/tensor transmission.
The impact of this can be seen by the blue thresholds in Figure~\ref{fig:clientdiff}. Compared to sending a fixed portion (red threshold) as in \cite{passban-etal-2022-training}, which can lead to sending too much (i.e., excessive communication) or too little (i.e., compromised NMT quality), our dynamic threshold accounts for varying NMT quality.}

\vspace{-2mm}
\subsection{Summary of Contributions}
Overall, we make the following major contributions in developing {\texttt{MetaSend}}:

\begin{itemize}
    \item We conduct the first research on the communication efficiency of FL in multilingual NMT, and study the relationship between translation quality and the volume of transmitted parameters in multilingual NMT engines.
    \item We propose a novel meta-learning-based adaptive parameter selection method to partition the tensors at a client into those which have vs. have not evolved significantly since the last FL transmission, balancing between translation quality and communication burden.
    \item Through extensive experiments on two benchmark NMT datasets with different language distributions, we demonstrate that our {\texttt{MetaSend}} methodology consistently enhances translation quality compared to baselines while operating within a fixed FL communication budget.
\end{itemize}
The rest of the paper is structured as follows. Section II reviews previous work on efficient methods and federated learning for NLP. Section III presents the problem setup of federated machine translation and introduces our proposed method, {\texttt{MetaSend}}, for enhancing efficiency. Section IV provides details about the experimental setups, while extensive experimental results are presented in Section V. Finally, we conclude our work in Section VI.

\vspace{-1mm}

\section{Related Work}

\subsection{Computational Efficiency in NLP}
Previous research has explored efficiency enhancements for large NLP models from a computational perspective, i.e.,  achieving comparable results with fewer resources~\cite{Treviso2022EfficientMF}. 
Some studies have focused on the data side, e.g., showing how smart downsampling of available datasets can result in equal or improved performance compared to using the entire dataset~\cite{Lee2021DeduplicatingTD,Zhang2022OPTOP}.
On the other hand, efforts to enhance efficiency through model designs include questioning the necessity of full attention heads in large language models and demonstrating that removing certain attention heads does not significantly impact test performance~\cite{Kovaleva2019RevealingTD,Raganato2020FixedES}. 
Several previous works have also investigated model pruning methods to improve the efficiency of learning large-scale language models, which identify and remove non-essential portions of networks.
Earlier work focused mainly on unstructured pruning, where weights are pruned individually~\cite{Narang2017ExploringSI, Zhu2017ToPO}. 
More recently, structured pruning removes groups of consecutive parameters or structured blocks of weights, leading to significant speedups~\cite{Li2020EfficientTL,Cao2019EfficientAE}. 
Compared to these works, motivated by the recent demand for FL in NLP, we focus on communication efficiency in federated multilingual NMT and design a strategy that selectively transmits only the essential parameters of NMT engines for learning.

\vspace{-2mm}

\subsection{Communication Efficiency in FL}
Communication efficiency is a critical aspect of FL, particularly in scenarios with limited bandwidth and resource-constrained devices (see~\cite{lu2024federated} for a recent survey). Extensive research has focused on enhancing communication efficiency through model and gradient compression techniques~\cite{Wang2022FederatedLW,Jeong2019MultihopFP}, alternative communication topologies~\cite{Parasnis2023ConnectivityAwareSF, lan2023fl}, and intelligent parameter selection and sparsification methods~\cite{ Wang2022FederatedLW, Chen2024SynchronizeOT}. 
Our work on selective parameter/tensor transmission is most relevant to the selection/sparsification category~\cite{passban-etal-2022-training,Wang2022FederatedLW,Chen2024SynchronizeOT}. 
Specifically, \cite{passban-etal-2022-training,Wang2022FederatedLW} sort the model parameters by absolute magnitude and select a subset of parameters that rank among the highest or lowest in the entire model, while \cite{Chen2024SynchronizeOT} uses historical model parameters to determine the important parameters for communication in the current model.
Different from these works, {\texttt{MetaSend}} is intended specifically for NMT tasks, which as discussed in Section~\ref{sec:intro} exhibit nuanced tensor deviations across training rounds due to language heterogeneity, regional variations, data imbalances, and other disparities across clients. Unlike these methods that consider only communication efficiency in their design, our method optimizes communication based on the model's performance, taking both performance and efficiency into account during the FL process.
We will see in Section~\ref{sec:exp} that our approach outperforms baselines~\cite{ Wang2022FederatedLW, Chen2024SynchronizeOT}, intended for more generic FL tasks, in the NMT setting we focus on.

\vspace{-2mm}

\subsection{Federated Learning and NLP}

Recent research has begun exploring FL methods for NLP applications requiring privacy preservation~\cite{Chu2022MitigatingBI,qin-etal-2021-improving-federated, Ge2020FedNERMN,lin-etal-2022-fednlp}. 
During the FL communication phase, large NLP models are exchanged, introducing a significant communication cost associated with model updates. 
To address this, \cite{MelasKyriazi2022IntrinsicGC} proposed a gradient compression methodology for language modeling, while \cite{ro-etal-2022-scaling} leveraged similarities between smaller and larger models in cross-device FL.
The parameter selection method proposed in \cite{passban-etal-2022-training} targeting mixed-domain NMT is most relevant to our work. 
As discussed in Section~\ref{sec:intro}, they analyze client deviations from the previous round and send a fixed portion of model tensors during FL communication. 
However, this fixed approach may overlook meaningful parameters, reducing translation quality. 
To address this, we design a meta-learning-based parameter selection method with dynamic thresholds to balance communication efficiency and translation quality.

\section{{MetaSend} for Federated NMT}

\subsection{Problem Setup: Federated NMT}\label{sec:Prob_setup}
When training the NMT model over multiple clients, we follow the general cross-silo FL setting introduced by~\cite{9599369}. Algorithm~\ref{alg:cap} summarizes the overall training procedure.
Each client first runs stochastic gradient descent (SGD) on its local data and then \texttt{Sends} the learned NMT model to the server.
The server then executes a global aggregation after receiving all the trained models. We assume the standard \texttt{FedAVG} algorithm~\cite{McMahan2017CommunicationEfficientLO} as the aggregation method. 
Using \texttt{FedAVG} as opposed to other aggregation procedures allow us to focus on the communication efficiency of NMT in the FL scenario. 

\begin{figure*}[t]
    \centering
    \setlength{\abovecaptionskip}{1mm}
\includegraphics[width=0.88\linewidth]{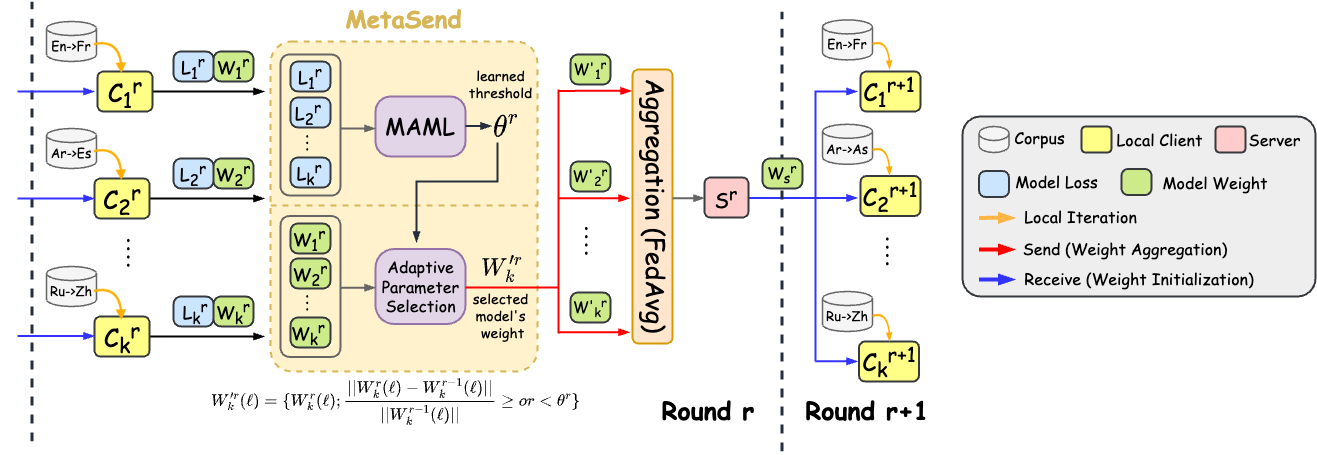}
    \caption{Overview of {\texttt{MetaSend}} for federated NMT. {\texttt{MetaSend}} allows clients to adaptively select key NMT model parameters using a learned threshold per round, sending only a subset of tensors to the server for aggregation and improving efficiency under a limited communication budget.}
    \label{fig:overview}
    \vspace{-3.5mm}
\end{figure*}

The \texttt{FedAVG} aggregation is defined as:
\begin{equation}\label{eq:agg}
W_s^r = \sum_{k=1}^{K} \frac{n_k}{n} W_k^r,
\end{equation}
where $K$ is the total number of clients, $n_k$ is the number of samples in the $k$-th client’s dataset, $n$ is the total number of all training data points, and $W_s^r$ and $W_k^r$ are the model parameters at the $r$-th communication round for server $\mathcal{S}^r$ and the $k$-th client $\mathcal{C}_k^r$, respectively.
The system has finished one FL communication round once the server has completed the aggregation.
For the next round, the clients will \texttt{Receive} the server’s weights for initialization. The overall process is repeated for $r = 1, 2, ..., R$ FL communication rounds.

\begin{algorithm}[t]
\caption{Cross-Silo Federated Learning}
\label{alg:cap}
\begin{algorithmic}[1]
\State Server $\mathcal{S}$, Client $\mathcal{C}_k$, total number of clients $K$
\For {Each round $r = 1, 2, ..., R$}
\State Each client \textbf{\texttt{Receive}}($\mathcal{S}^{r-1}$)
\State $\mathcal{C}_k^r$ $\gets$ \textbf{Local Iterations}
\State \textbf{\texttt{Send}}($\mathcal{C}_k^r$) to server
\State $\mathcal{S}^r$ $\gets$ \textbf{Aggregation}($\mathcal{C}_1^r, \mathcal{C}_2^r, ..., \mathcal{C}_K^r$)

\EndFor
\end{algorithmic}
\end{algorithm}

\vspace{-1mm}
\subsection{Overview of {MetaSend}}
\label{sec:overview_metaprune}
During FL communication, a large amount of NMT model weights have to be uploaded to the server during the \texttt{Send} action (in Algorithm~\ref{alg:cap}) for aggregation. 
This communication can be quite costly for a large NMT model, which is a key bottleneck for FL.
To tackle this challenge, we propose {\texttt{MetaSend}}, which adapts the NMT tensors sent based on a customized sending threshold for each communication round. 
The key idea of {\texttt{MetaSend}} is to build Model-Agnostic Meta-Learning (MAML)~\cite{Finn2017ModelAgnosticMF} into the FL rounds to balance communication efficiency and translation quality.

Figure~\ref{fig:overview} and Algorithm~\ref{alg:flmaml} summarize the overall procedure. In each round $r$, after completing the local iterations using their local training data, each client $\mathcal{C}^r_k$ will retain its learned model weight $W^r_k$ and training loss $L^r_k$. 
In each round $r$, {\texttt{MetaSend}} operates according to the following steps:
(i) After every client has finished training their local models, the training losses $L^r_1, L^r_2, ..., L^r_K$ of all $K$ clients are inputted into our MAML module, which is implemented as a multi-layer perceptron (MLP) network. The MAML module serves as a server-side component that leverages the clients' losses to learn a threshold, which is subsequently shared with all clients.
The purpose of this module is to generate a customized threshold $\theta^r$ based on the extracted losses (line 5 in Algorithm~\ref{alg:flmaml}), which should consider the anticipated impact on learning performance.
(ii) Based on the threshold $\theta^r$, each client $\mathcal{C}^r_k$ selects which model tensors to send based on a deviation comparison with its previous version $\mathcal{C}^{r-1}_k$ (line 6 in Algorithm~\ref{alg:flmaml}). 
(iii) After receiving the transmissions, the server $\mathcal{S}^r$ executes the aggregation by taking the resulting models' weights from all clients (line 8 in Algorithm~\ref{alg:flmaml}).
(iv) Subsequently, the MAML module is updated through meta-learning (line 9 in Algorithm~\ref{alg:flmaml}), taking into account the translation quality of the global model at the server $\mathcal{S}^r$.

There are two key challenges in designing {\texttt{MetaSend}}: (a) How to design the sending criterion for NMT models in Step (ii)? (Answered in Section~\ref{sec:pruneagg}); (b) How can the MAML module effectively learn to produce a customized threshold for each FL round in Step (iv)? (Answered in Section~\ref{sec:maml})



\vspace{-2mm}

\subsection{{MetaSend}: Customized Sending and Aggregation}
\label{sec:pruneagg}

In this subsection, we answer the first question mentioned above. 
Our intuition is that the extent of parameter deviation relative to the original norm provides an indication of whether information is worth sending. 
Our observation in Section~\ref{sec:intro} shows that the tensors of the NMT model responsible for learning exhibit a clustered pattern in the deviation distribution.

Compared with the clients in the previous round, {\texttt{MetaSend}} will first compute the deviation ($dev$) for each tensor, with $dev$ defined as:
\begin{equation} 
dev = \dfrac{||W^{r}_{k}(\ell)-W^{r-1}_{k}(\ell)||}{||W^{r-1}_{k}(\ell)||},
\end{equation}
where $\ell \in \mathcal{L}$ denotes a particular tensor of the model, and $||\cdot||$ is the absolute-value norm that measures the difference between clients' weights in different rounds.
Based on $dev$ and the learned threshold $\theta^r$ (line 6 in Algorithm~\ref{alg:flmaml}), {\texttt{MetaSend}} may select each tensor to be sent based on one of two criteria: whether its $dev$ is greater ($g$) or less ($l$) than the threshold $\theta^r$. 
Each of these has potential advantages: deviations above the threshold ($g$) will promote sending tensors that have experienced the largest changes, which could potentially be an informative or noisy update, while deviations below the threshold ($l$) will encourage more gradual tensor refinements that are not susceptible to sudden large fluctuations.
As a result, {\texttt{MetaSend}} generates two sending methods, namely $\text{\texttt{MetaSend}}_g$ and $\text{\texttt{MetaSend}}_l$:
\begin{equation}
\label{eq:metaprune}
\left\{
\begin{aligned}
\text{\texttt{MetaSend}}_g: W'^{r}_{k}(\ell)= \{W^{r}_{k}(\ell); dev \geq \theta^r\},\\
\text{\texttt{MetaSend}}_l:W'^{r}_{k}(\ell)=
\{W^{r}_{k}(\ell); dev < \theta^r\}, \\
\end{aligned}
\right.
\end{equation}
where $W'^{r}_{k}$ represents the selected model's weights for the $k$-th client in round $r$.

Given the resulting weights $W'^{r}_{k}$ of every client, the server then executes aggregation via \texttt{FedAVG} (line 8 in Algorithm~\ref{alg:flmaml}). Formally, Equation~\ref{eq:agg} can be re-formulated as:
\begin{equation}\label{eq:reagg}
W_{s}^{r} = \sum_{k=1}^{K} \frac{n_k}{n} \left(W'_{k}\right)^{r}.
\end{equation}

\subsection{{MetaSend}: MAML Module Update}
\label{sec:maml}


\begin{algorithm}[!t]
\caption{FL with MetaSend}
\label{alg:flmaml}
\begin{algorithmic}[1]
\State Model Parameter: $W_s$ for server $\mathcal{S}$, $W_k$ for each client $\mathcal{C}_k$ , $\phi$ for MAML module (MLP)
\For {Each round $r = 1, 2, ..., R$}
\State Initialize all $K$ clients by $W_{s}^{r-1}$
\State $W_{k}^{r}$, $L_{train}(W_{k}^{r})$ $\gets$ \textbf{Local Iterations}
\State	$\theta^r$ $\gets$ \textbf{MLP}($L_{train}(W_{1}^{r}),..., L_{train}(W_{K}^{r})$)
\State ${W'_{k}}^{r}$ $\gets$ \textbf{MetaSend}($W_{k}^{r}$, $\theta^r$)
\State Send resulting local model${W'_{k}}^{r}$ to server
\State $W_{s}^{r}$ $\gets$ \textbf{Aggregation}(${W'_{1}}^{r}, ..., {W'_{k}}^{r}$)
\State $\phi^{r}$ $\gets$ \textbf{MetaUpdate}($L_{val}(W^{r}_{s}$))
\EndFor 
\end{algorithmic}
\end{algorithm}

\begin{figure}
    \centering
    \setlength{\abovecaptionskip}{1mm}
    \includegraphics[width=0.9\linewidth]{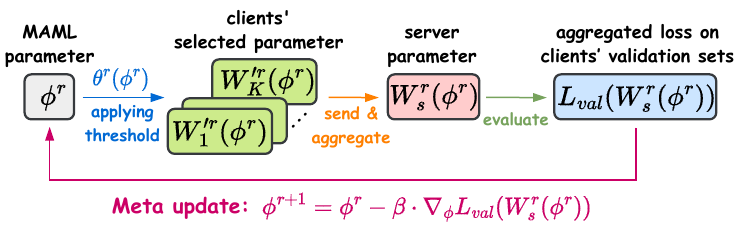}
    \caption{Optimization of our MAML module in an FL setup involves adapting the sending threshold based on NMT model quality. The threshold is initially applied to every client, resulting in selected parameters for each client. Subsequently, clients send the resulting parameters to the server for aggregating a global model. After aggregating the client models, we evaluate the global model using validation sets from each client. Finally, the MAML module performs a meta-update based on the evaluated loss.}
    \label{fig:MAML}
    \vspace{-3mm}

\end{figure}
To address the second question, our MAML module generates an adaptive sending threshold based on translation quality, as illustrated in Figure~\ref{fig:MAML}.
$\phi^r$ represents the hyperparameter set of our MAML module in communication round $r$, and $\theta^r(\phi^r)$ is the generated threshold from the module.
After the parameter selection process and aggregation (Equations~\ref{eq:metaprune} and \ref{eq:reagg}), the parameters of the $k$-th resulting client model and the global model can be expressed as $W'^r_k(\phi^r)$ and $W^r_s(\phi^r)$, respectively. 
To assess the quality of the global model $W^r_s(\phi^r)$, we randomly select $b$ batches of samples from the validation dataset and evaluate the global model using these samples. 
Subsequently, we employ the validation loss $L_{val}(W^r_s(\phi^r))$ as the MAML module's optimization objective to guide it toward improving translation quality.
Thus, our MAML module update can be written as: 
\begin{equation}\label{eq:maml}
\phi^{r+1} =\phi^r-\beta\cdot\nabla_{\phi}L_{val}(W^r_s(\phi^r)),
\end{equation}
where $\beta$ is the learning rate for the meta update.
It is important to note that although we are evaluating the global model, the validation loss is actually calculated using the validation dataset present in each client. 
This process can be seen as occurring after broadcasting the global model to each client at every round. In this process, each client computes the validation loss using the global model and sends the validation loss back to the server. Subsequently, the server updates the MAML module according to the aggregated loss.
Finally, by optimizing the MAML module with consideration of the translation quality of the global NMT model, our MAML module can generate a customized threshold $\theta^r$ for each round that considers both the deviation distribution and the translation quality.
We will see in Section~\ref{sec:exp} how this process of learning what parameters to send results in substantial translation quality and communication efficiency improvements.


\section{Experimental Design}
\label{sec:exp_design}
\subsection{Datasets and Client Partitioning}
We utilize two widely used NMT datasets: Machine Translation of Noisy (MTNT) dataset~\cite{michel-neubig-2018-mtnt} and UN Machine Translation (UNMT) Corpus~\cite{ziemski-etal-2016-united}.
The MTNT dataset was gathered from user comments on Reddit. 
The dataset contains two language directions: English to French (En $\to$ Fr) and English to Japanese (En $\to$ Ja).
The dataset contains 5,605 instances in each direction for training and 1k each for validation and test set. {MTNT features diverse language pairs, each with unique linguistic characteristics, making it a strong candidate for testing the ability of our methodology to handle various grammatical, syntactic, and semantic challenges.}
The UNMT Corpus consists of manually translated UN documents, and we consider three language directions: English to French (En $\to$ Fr), Arabic to Spanish (Ar $\to$ Es), and Russian to Chinese (Ru $\to$ Zh).
The dataset contains 80k instances in each direction for training and 10k each for validation and test set. Similar to MTNT, UNMT captures a wide range of linguistic variations across its six languages, each with unique grammar, syntax, and semantic structures.
For each dataset, we consider three training settings: (i) centralized training without FL, (ii) FL with IID data, where the data for each client is sampled randomly from all data, and (iii) FL with non-IID data, where each client only sees data for one language direction. 





\begin{table*}
\caption{SacreBLEU scores obtained with centralized and FL (IID and Non-IID) methods for various strategies on the MTNT and UNMT datasets. The bold scores indicate that \texttt{MetaSend} outperforms other methods in all cases.}

\small
\begin{center}
\begin{tabular}{ccccccccc}
\hline
\multirow{2}{*}{\textbf{Training}}& \multirow{2}{*}{\textbf{Method}}& \multicolumn{3}{c}{\textbf{MTNT}}  & \multicolumn{4}{c}{\textbf{UNMT}}  \\ 
\cmidrule(l){3-5}\cmidrule(l){6-9}
& & \textbf{En $\to$ Fr} & \textbf{En $\to$ Ja} & \textbf{Avg} & \textbf{En $\to$ Fr} & \textbf{Ar $\to$ Es} & \textbf{Ru $\to$ Zh} & \textbf{Avg}\\
\hline\hline

{Centralized}& w/ pre-trained & 33.4 & 21.7 & 27.6 & 44.5& 44.5& 45.0 & 44.7\\
IID FL & PMFL & 32.3 & 17.9 & 25.1 & 43.7 & 39.2 & 42.1 & 41.7\\\
Non-IID FL & PMFL & 20.6 & 13.9 & 17.2 & 30.1 & 30.8 & 29.4 & 30.1 \\
\hline
\multirow{5}{*}{IID FL}& RandSend & 29.7 & 11.4 & 20.6 & 35.2 & 33.7 & 36.9 & 35.3\\
& DP$_{g}$ & 31.1 & 14.6 & 22.9 & 37.9 & 38.5 & 38.7 & 38.4\\	
& DP$_{l}$ & 31.9	& 15.5 & 23.7 & 38.4 & 38.7 & 39.2 & 38.8\\

& {MetaSend}$_g$ (ours)& 32.0 & 16.8 & 24.4 & 42.0 & 39.1 & \textbf{41.9} & 41.0\\
& {MetaSend}$_l$ (ours)& \textbf{32.7} & \textbf{17.3} & \textbf{25.0} & \textbf{42.3} & \textbf{39.3} & 41.6 & \textbf{41.1} \\
\hline
\multirow{5}{*}{Non-IID FL}& RandSend & 16.1 & 10.0 & 13.1 & 22.6 & 24.3 & 16.8 & 21.2\\
& DP$_{g}$ & 19.0 & 12.6 & 15.8 & 29.8 & 28.3 & 	26.5 & 28.2
\\	
& DP$_{l}$ & 18.3 & 13.1 & 15.7 & 30.5 & 29.9 &	26.1 &	28.8
\\
& {MetaSend}$_g$ (ours)& 19.7 & \textbf{13.9} & 16.8 & 31.2 & 30.1 & \textbf{29.7} & 30.3\\
& {MetaSend}$_l$ (ours)& \textbf{20.1} & 13.6 & \textbf{16.9} & \textbf{32.8} & \textbf{31.3} & \textbf{29.7} & \textbf{31.3}\\
\hline
\end{tabular}
\label{tb:bleu_score}
\vspace{-4mm}
\end{center}
\end{table*}

\begin{figure}[t]
    \centering
    \setlength{\abovecaptionskip}{1mm}
    \includegraphics[width=0.9\linewidth]{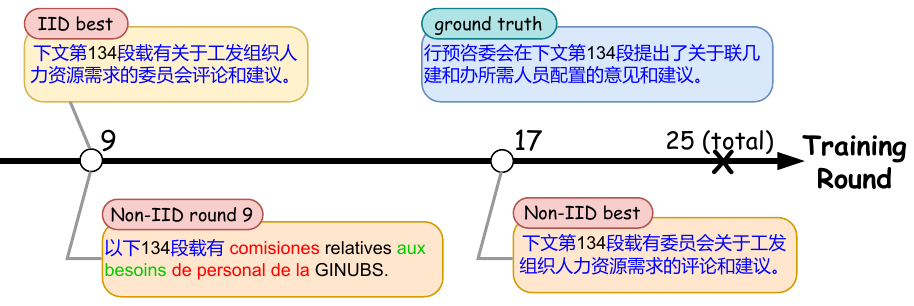}
    \caption{The generations (Ru $\to$ Zh) of ${\text{\texttt{MetaSend}}_{l}}$ under IID and Non-IID FL training. Similar to most FL methods, models under Non-IID FL training need more training rounds to reach stability. The generations before convergence consists of different languages for the early model in Non-IID FL training. ({\color{blue}blue}: Zh, {\color{red}red}: Es, {\color{green}green}: Fr) 
    }
    \label{fig:converg}
    \vspace{-3.5mm}
\end{figure}

\subsection{Base Model and Evaluation Metrics}
Following the multilingual FL experimental settings in ~\cite{weller-etal-2022-pretrained}, we use the M2M-100 model~\cite{Fan2021BeyondEM} to conduct machine translation.
The M2M-100 model is a sequence-to-sequence model with 418M parameters, and it can translate between any pair of 100 languages.
We measure the quality of translation by using sacreBLEU~\cite{Post2018ACF} and COMET~\cite{rei-etal-2020-comet}.
SacreBLEU is a commonly used metric for evaluating NMT quality, while several research works have mentioned that the traditional overlap-based evaluation metrics do not correlate well with human evaluation~\cite{sinha-etal-2020-learning,Hsu2022LearningTR, Hsu2021PlotAR}.
Therefore, we also evaluated each method using a more advanced MT evaluation metric, COMET, demonstrating that the evaluated quality is better aligned with human judgement.
Besides translation quality, we consider two metrics for FL efficiency: tensor saving and processing time. Tensor saving is defined as the ratio of tensors that are not exchanged between the server and clients during the \texttt{Sending} step in Algorithm~\ref{alg:cap} (or line 7 in Algorithm~\ref{alg:flmaml}). For efficiency evaluation, we will report the average tensor savings and the exact processing time over all training rounds.



\subsection{FL Training and MAML Module}

We build our FL experiments using the Flower framework~\cite{Beutel2020FlowerAF} for training and evaluation.
For centralized experiments, we train  models for 50 epochs and discuss the effect of pre-trained knowledge for NMT.
For every FL experiment, we train each method for 25 communication rounds (epochs) and initialize the clients using a pre-trained M2M-100 model from Hugging Face's transformers library~\cite{Wolf2019HuggingFacesTS}. As a reference, we also conduct FL experiments by initializing the clients' model with random weights.


For our MAML module, we use a multi-layer perceptron (MLP) network with one hidden layer containing 100 neurons as the default setting~\cite{Chu2021ClickBasedSP}.
The ablation study in Section~\ref{sec:ablation_neuron} presents the results of {\texttt{MetaSend}} considering different numbers of neurons in the MAML module. 
To randomly sample a small portion of the validation set for the MAML update, we use 16 batches (i.e., $b=16$). 
In Section~\ref{sec:ablation_batch}, we also present an ablation study on $b$ to see the effect of MAML optimization for NMT quality. 




\vspace{-2mm}

\subsection{Baselines}
We use several competitive baseline approaches and parameter selection strategies compare with {\tt MetaSend} in federated NMT.
\textbf{PMFL}~\cite{weller-etal-2022-pretrained} is the basic FL framework that uses a pre-trained model for federated NMT without any decision-making mechanism.
\textbf{${\text{DP}_{g}}$} and \textbf{${\text{DP}_{l}}$} are recent methods from~\cite{passban-etal-2022-training} that select which tensors to send by comparing the norm difference between the previous and current client models. 
Their thresholding mechanism first sorts the parameters by norm differences and either sends the top 50\% ({${\text{DP}_{g}}$) or the bottom 50\% ({${\text{DP}_{l}}$) of parameters for aggregation.
We also include the results from a random configuration, \textbf{RandSend}, which randomly sends 50\% of the tensors during FL aggregation.
In addition to the baselines designed for federated multilingual NMT,we compare our method to \textbf{top-k sparsification}~\cite{Wang2022FederatedLW} and \textbf{Adaptive Parameter Freezing (APF)}~\cite{Chen2024SynchronizeOT}, two recent communication-efficient FL methods intended for more generic ML tasks. 
Top-k sparsification sorts the model parameters based on their absolute values and sends only the largest k parameters for aggregation.
APF employs an observation window consisting of a certain number of consecutive model updates from recent past rounds to compute perturbations and determine which parameters need to be communicated. 
}


\vspace{-1mm}

\section{Experimental Results}
\label{sec:exp}

\subsection{Translation Performance Evaluation}

\label{sec:result_translation}

\begin{table*}
\caption{COMET scores obtained with centralized and different FL methods on MTNT and UNMT datasets. In addition to traditional evaluation metrics, the improved COMET scores show that our method provides translations that align with human preference.}

\small
\begin{center}
\begin{tabular}{ccccccccc}
\hline
\multirow{2}{*}{\textbf{Training}}& \multirow{2}{*}{\textbf{Method}}& \multicolumn{3}{c}{\textbf{MTNT}}  & \multicolumn{4}{c}{\textbf{UNMT}}  \\ 
\cmidrule(l){3-5}\cmidrule(l){6-9}
& & \textbf{En $\to$ Fr} & \textbf{En $\to$ Ja} & \textbf{Avg} & \textbf{En $\to$ Fr} & \textbf{Ar $\to$ Es} & \textbf{Ru $\to$ Zh} & \textbf{Avg}\\
\hline\hline

{Centralized}& w/ pre-trained & 0.778 & 0.759 & 0.769 & 0.875 & 0.863 & 0.855	& 0.864\\
{IID FL}& PMFL & 0.758 & 0.734 & 0.746 & 0.853 & 0.839 & 0.830 & 0.841\\
{Non-IID FL}& PMFL & 0.666 &{ 0.659} & {0.663} & 0.737 & 0.742 & 0.709 & 0.729\\ 
\hline
\multirow{5}{*}{IID FL}& RandSend & 0.726 & 0.715 & 0.721 & 0.811 & 0.773 & 0.811 & 0.798\\
& DP$_{g}$ & 0.729 & 0.721 &	0.725 &	0.825	& 0.801 & 0.823 &	0.816\\	
& DP$_{l}$& 0.737 & 	0.730 & 0.734 & 0.830 & 	0.805 & 0.829 & 0.821\\
& {MetaSend}$_g$ (ours)& \textbf{0.755} & 0.733 & 0.744 & 0.846 & \textbf{0.829} & \textbf{0.833} & \textbf{0.836} \\
& {MetaSend}$_l$ (ours)& \textbf{0.755} & \textbf{0.736} & \textbf{0.746} & \textbf{0.849} & 0.827 & 0.831 & \textbf{0.836}\\
\hline
\multirow{5}{*}{Non-IID FL}& RandSend & 0.639 & 0.623 & 0.631 & 0.651 & 0.663 & 0.659 & 0.658\\
& DP$_{g}$ & 0.651 & 0.650 & 0.651 & 0.709 & 0.720  & 0.684 & 0.704\\	
& DP$_{l}$ & 0.659 & 0.648 & 0.653 & 0.713 & 0.729 & 0.692 & 0.711\\
& {MetaSend}$_g$ (ours)& \textbf{0.668} & 0.653 & 0.661 & 0.744 & 0.739 & 0.713 & 0.732\\
& {MetaSend}$_l$ (ours)& 0.666 & \textbf{0.659} & \textbf{0.663} & \textbf{0.749} & \textbf{0.744} & \textbf{0.721} & \textbf{0.738}\\
\hline
\end{tabular}
\label{tb:comet_score}
\vspace{-3.5mm}
\end{center}
\end{table*}

\begin{figure*}
    \centering
    \setlength{\abovecaptionskip}{1mm}
    \includegraphics[width=1.0\linewidth]{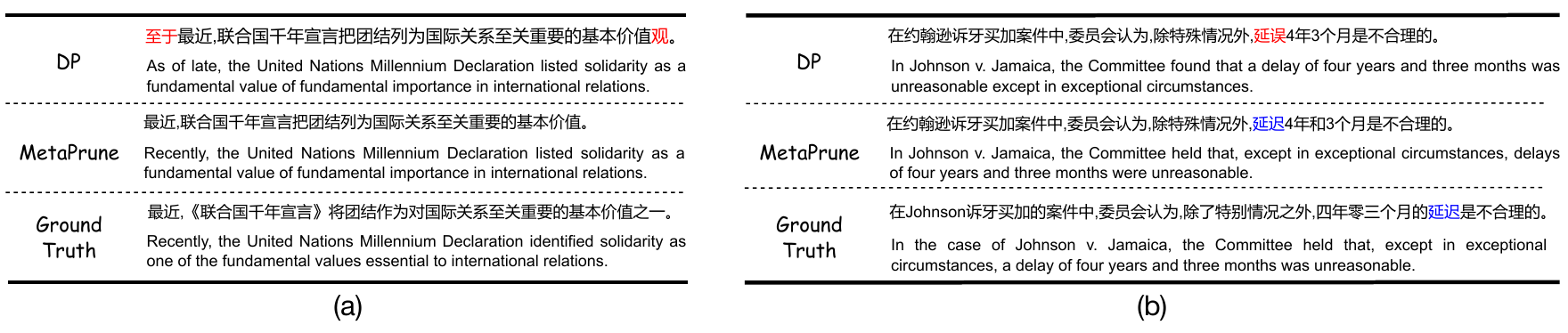}
    \caption{(a) Translation examples (Ru $\to$ Zh) of ${\text{DP}_{l}}$, ${\text{\texttt{MetaSend}}_{l}}$, and ground truth. Our method aligns better with ground truth, and ${\text{DP}_{l}}$ generates redundant tokens. (b) Translation  examples (Ru $\to$ Zh) of ${\text{DP}_{l}}$, ${\text{\texttt{MetaSend}}_{l}}$, and ground truth. Our method generates the same sentiment-meaning word as ground truth, while ${\text{DP}_{l}}$ generates similar but different sentiment-meaning words.}
    \label{fig:examples}
    \vspace{-3.5mm}
\end{figure*}
Table~\ref{tb:bleu_score} and ~\ref{tb:comet_score} present the SacreBLEU and COMET results of the translation task for both MTNT and UNMT datasets. 
In the first section of the tables, we observe that the centralized method outperforms PMFL methods, as we would expect; on the other hand, it compromises data privacy by not preserving individual client data confidentiality.
Further, the performance decrease of every method from IID to non-IID FL training reveals the challenges in the practical NMT scenario of clients having only single language directions.
Figure~\ref{fig:converg} shows the challenge of training FL algorithm under Non-IID data distribution.
We can see that our method, trained under the IID setting, achieves a good translation compared to the ground truth by round 9. 
However, when trained under the Non-IID setting, our method requires 17 communication rounds to achieve comparable results. 
The result obtained at round 9 for our method trained under the Non-IID setting shows that the model has not yet converged, resulting in chaotic translations with mixed languages.

By randomly sending the model parameters, RandSend achieves the lowest performance among all methods for both IID and non-IID FL. 
Compared with our MetaSend methods, we see that the DP methods face challenges.
The significant performance improvements of {\texttt{MetaSend}} over DP show the advantage of modeling an adaptive sending threshold based on the norm difference distribution. 
Moreover, our MAML-learned threshold learns what to send during communication to better optimize the NMT task. 
Note that this threshold is dynamic and can adapt to different norm difference distributions in each round.
Specifically, our {\texttt{MetaSend}} method achieves average sacreBLEU improvements of 3.9 and 3.4 points over DP on IID and non-IID data, respectively.

Among all the parameter selection methods, {$\text{\texttt{MetaSend}}_l$} achieves the highest scores in both sacreBLEU and COMET metrics. 
It demonstrates comparable translation quality to PMFL, indicating its ability to preserve communication resources without compromising translation quality.
Translation examples generated by our method and the baseline are provided in Figure~\ref{fig:examples}, where it is evident that our method shows better alignment with the ground truth regarding sentiment and accurate word usage. 
Figure~\ref{fig:examples}(a) illustrates that our method generates translations that closely align with the ground truth by utilizing similar words. In contrast, the baseline method produces translations with redundant tokens, leading to potential confusion within the sentence.
Figure~\ref{fig:examples}(b) shows that our method employs the same words as the ground truth, conveying a neutral sentiment. In comparison, the baseline method generates similar words but with a negative sentiment.

{In addition to the baselines DP$_{l}$ and DP$_{g}$ that are explicitly designed for federated NMT scenarios, we also  evaluate the performance of top-$k$ sparsification and APF in this language domain. Table~\ref{tb:rebuttal_apf} presents the average translation results along with the parameter savings obtained by our method and these baselines across both datasets and training settings. 
The value of $k$ is set according to 50\% of the largest model parameters for top-$k$ compression, and the historical window size for APF is set to 4. 
The results demonstrate that our {MetaSend}$_l$ achieves the highest average translation quality, while our {MetaSend}$_g$ consistently outperforms top-$k$ compresion and APF in both translation performance and communication efficiency. 
This highlights the effectiveness of our MAML-based adaptive thresholding technique, which is designed to address the nuances of federated multilingual NMT. 
Moreover, while these parameter-wise methods focus on communication efficiency, they fail to account for the unique deviation patterns in NMT, which can result in sending either too many or too few parameters, missing opportunities for improving either communication efficiency or translation performance without noticeably affecting the other metric. The thresholds reflected by these approaches are based solely on parameter magnitude or past parameters and are not optimized with consideration for the model's performance in federated NMT, as our method achieves. These differences highlight the advantages of our approach in balancing communication efficiency and maintaining model performance for multilingual NMT tasks.
}

\begin{table*}
\caption{Translation performance and communication efficiency comparison with other communication-efficient FL methods.}

\small
\begin{center}
\begin{tabular}{cccccc}
\hline
\multirow{2}{*}{\textbf{Training}}& \multirow{2}{*}{\textbf{Method}}& \multicolumn{2}{c}{\textbf{MTNT}}  & \multicolumn{2}{c}{\textbf{UNMT}}  \\ 
\cmidrule(l){3-4}\cmidrule(l){5-6}
& & \textbf{Translation Performance} & \textbf{Tensor Saving} & \textbf{Translation Performance} & \textbf{Tensor Saving} \\
\hline

\multirow{4}{*}{IID FL} & top-k compression  & 23.0 & 50.0\% & 38.2 & 50.1\%  \\
& APF & 23.9 & 58.7\% & 41.0 & 52.7\% \\
& {MetaSend}$_g$ (ours) & 24.4 & 61.5\%& 41.0 & 59.2\%  \\
& {MetaSend}$_l$ (ours)&  25.0 & 41.9\% & 41.1 & 44.5\%\\
\hline

\multirow{4}{*}{Non-IID FL} & top-k compression & 15.7 & 49.9\% & 28.2 & 49.9\%  \\
& APF & 16.4 & 49.3\% & 29.4 & 41.2\% \\
& {MetaSend}$_g$ (ours) & 16.8 & {62.3\%} & 30.3 & {58.3\%}\\
& {MetaSend}$_l$ (ours)&  {16.9} &42.7\% & {31.3} & {43.9\%} \\
					
\hline
\end{tabular}
\label{tb:rebuttal_apf}
\end{center}
\end{table*}

\subsection{Communication Efficiency Evaluation}

\begin{figure*}[t]
    \centering
    \includegraphics[width=0.9\linewidth]{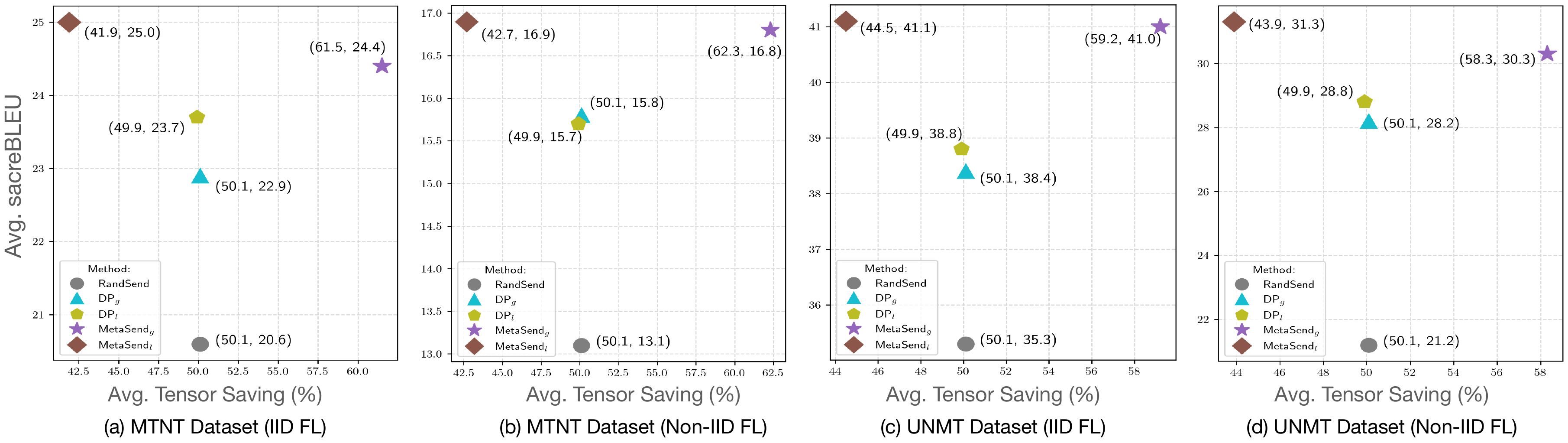}
    \caption{The average sacreBLEU score and tensor savings on MTNT and UNMT datasets show that {\texttt{MetaSend}} outperforms baselines in translation quality and communication efficiency, emphasizing the value of a dynamic threshold accounting for translation quality and tensor distribution.}
    \label{fig:efficient}
    \vspace{-3.5mm}
\end{figure*} 

\begin{table*}[!ht]
\caption{Average training time (in seconds) over 25 rounds for each method. Local training times ($\checkmark$) are similar across methods, as all require full parameter training. Overall, we observe that our method performs faster compared to other baselines.
}
\small
\begin{center}
\begin{tabular}{cccccc|ccc}
\hline
\textbf{Method}   & \begin{tabular}[c]{@{}c@{}} \textbf{Local} \\ \textbf{Training}\end{tabular} & \begin{tabular}[c]{@{}c@{}} \textbf{Compare Layers} \\ \textbf{\& Select}\end{tabular} & \begin{tabular}[c]{@{}c@{}} \textbf{MAML} \\ \textbf{Module}\end{tabular}  & \begin{tabular}[c]{@{}c@{}} \textbf{Send} \\ \textbf{Clients}\end{tabular}& \multicolumn{1}{c|}{\textbf{Aggregation}} & \begin{tabular}[c]{@{}c@{}} \textbf{Total Time} \\ \textbf{(w/o Local Training)}\end{tabular}
\\ 
\hline\hline

PMFL & $\checkmark$ & --- & --- & 32.712 & 18.224 & 50.936 \\
DP$_g$ & $\checkmark$ & 5.492  & --- & 22.359 &  11.195 & 39.046\\
DP$_l$ & $\checkmark$ & 5.510 & --- & 23.107 & 12.371 & 40.988\\ 
{MetaSend}$_g$ & $\checkmark$ & 3.547 & 0.878 & 21.395 & 10.795 & \textbf{36.615}\\
{MetaSend}$_l$  & $\checkmark$ & 3.485 & 0.893 & 23.519 & 12.914 & 40.811\\

\hline
\end{tabular}

\end{center}

\vspace{-1.0pc}
\label{tb:time}
\end{table*}


In addition to translation quality, recall that one of our key objectives is improving efficiency during FL communication.
In Figure~\ref{fig:efficient}, we present the average sacreBLEU score and tensor savings for each method across 25 communication rounds on the MTNT (a and b) and UNMT (c and d) datasets. 
The RandSend and DP methods can save around 50\% of tensors during FL communication due to their designs.
By sending the tensors based on a specific learned threshold, {\texttt{MetaSend}} methods obtain substantial improvements in translation quality compared to DP methods, while also obtaining varying degrees of tensor savings.
Among our two {\texttt{MetaSend}} methods, {\texttt{MetaSend}}$_l$ outperforms in translation quality, indicating that sending the majority of tensors for update ensures significant performance improvement. 
On the other hand, {\texttt{MetaSend}}$_g$ demonstrates higher tensor savings, with an average of 10.3\% more tensors saved compared to DP methods. 

{Figure~\ref{fig:saving_per_round} visualizes the translation quality and parameter savings obtained over federated training rounds for DP and our method, under the non-IID FL scenario. This offers a more detailed perspective on performance and communication efficiency throughout the training process. We can see that {\texttt{MetaSend}}$_g$ consistently provides efficiency benefits over the DP method across all rounds. Recall from Figure~\ref{fig:clientdiff} that tensor distributions over FL rounds tend to cluster predominantly in low-variation regions, with some tensors falling into highly active areas. {\texttt{MetaSend}}$_g$, which selects parameters greater than the dynamically learned threshold, often transmits fewer parameters, resulting in higher tensor savings across rounds. In contrast, {\texttt{MetaSend}}$_l$, which selects parameters below the threshold, typically transmits more parameters, leading to lower tensor savings. These differences highlight how the dynamic thresholds in {\texttt{MetaSend}} adapt to varying tensor deviations, determining crucial parameters for communication based on the machine translation performance.
}

\begin{figure*}
    \centering
    \setlength{\abovecaptionskip}{1mm}
    \includegraphics[width=0.85\linewidth]{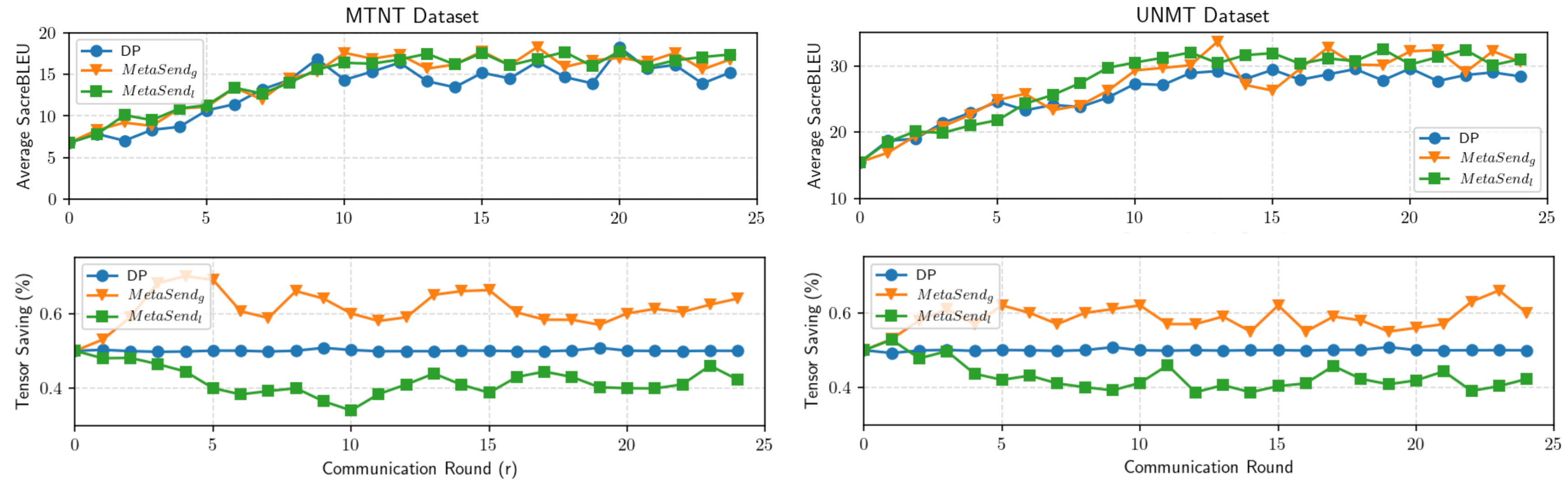}
    \caption{Translation performance and tensor savings across training rounds under the non-IID FL scenario for the MTNT and UNMT datasets.}
    \label{fig:saving_per_round}
    \vspace{-3.5mm}

\end{figure*}


In addition to evaluating tensor savings, Table~\ref{tb:time} reports training times for each method.  
All methods require similar time for local iterations, processing the full dataset and updating all parameters.
While PMFL skips parameter selection, it spends the most time on communication and aggregation as it transmits and aggregates all tensors. 
Both DP and {\texttt{MetaSend}} compute deviations between a client's current and previous round tensors. 
While DP sorts all deviations and selects the top or bottom 50\% of tensors, {\texttt{MetaSend}} uses a learned threshold to decide transmissions without sorting, resulting in faster parameter selection.
Additionally, {\texttt{MetaSend}}$_g$, which transmits the fewest parameters, requires the least time for sending and aggregation.

Both {\texttt{MetaSend}}$_l$ and {\texttt{MetaSend}}$_g$ involve additional computation for the MAML module, and they spend a similar amount of time on this module as it is independent of the operator. 
Table~\ref{tb:time_maml} provides a breakdown of time spent on the MAML module for {\texttt{MetaSend}}.
We observe that meta-evaluation is the most time-consuming, involving forward passes to the global model and computing the validation loss.
Therefore, we conduct an ablation study in Section~\ref{sec:ablation_batch} to examine the impact of the number of sampled batches.
In sum, our proposed {\texttt{MetaSend}} significantly enhances translation quality while achieving greater resource savings compared to both PMFL and DP methods.

\begin{table}
\caption{Time analysis of our MAML module shows meta-evaluation is the most time-consuming operation. We perform an ablation study in Section~\ref{sec:ablation_study} to examine its trade-offs.}

\small
\begin{center}
\begin{tabular}{cc}
\hline
\textbf{Operation}  & \textbf{Time (sec)} \\
\hline\hline

Forward MLP (size:100) \& Ouput $\theta^r$ & 7.275 $\times 10^{-4}$  \\
Meta Evaluation (16 batches) & 8.754 $\times 10^{-1}$ \\
Meta Optimization & 1.137 $\times 10^{-3}$\\

\hline
\end{tabular}

\end{center}
\label{tb:time_maml}
\vspace{-3.5mm}
\end{table}

\begin{figure*}[t]
    \centering
    \includegraphics[width=0.95\linewidth]{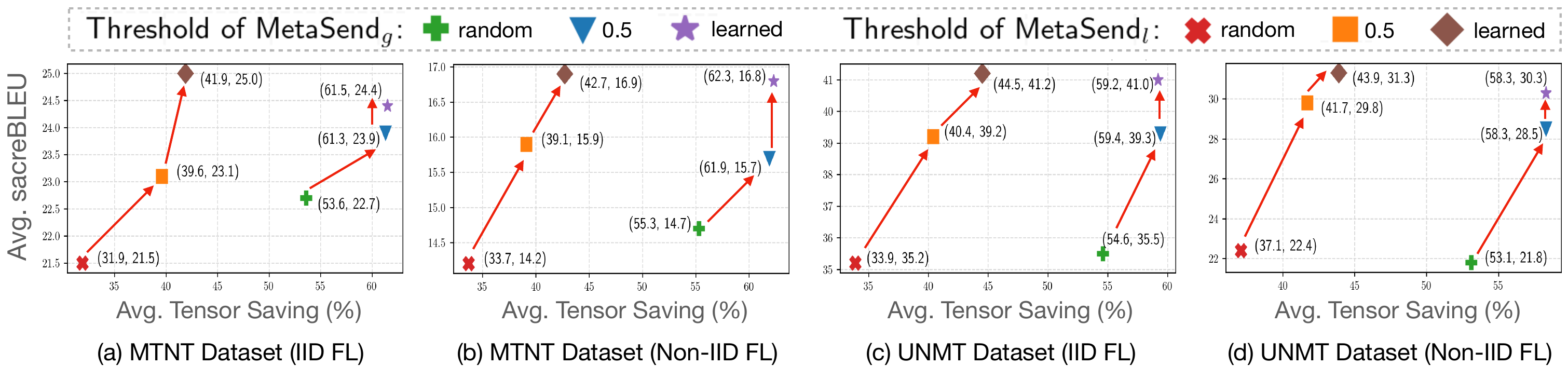}
    \caption{Average sacreBLEU score and tensor savings of {\texttt{MetaSend}} with random, fixed, and our learned sending thresholds. Our learned sending threshold consistently demonstrates improvements in both translation performance and efficiency, highlighting the significance of a dynamically learned sending threshold.}
    \label{fig:ablation}
\end{figure*}

\color{black}
\subsection{Varying System Parameters and Configuration}
\label{sec:ablation_study}

\subsubsection{Effectiveness of learned threshold} 
\label{sec:ablation_threshold}
To isolate the impact of the sending threshold, we compare {\texttt{MetaSend}} with different thresholds, including our learned threshold, a fixed threshold ($\theta^r = 0.5$), and a random threshold selected from 0 to 1.
The red arrows in Figure~\ref{fig:ablation} show the improvements in {\texttt{MetaSend}} when using different thresholds within a single operator ($l$ or $g$). 
Compared with other thresholds, our learned threshold improves both translation quality and tensor savings.
By comparing Figures~\ref{fig:ablation} and~\ref{fig:efficient}, {\texttt{MetaSend}} with a fixed threshold sometimes outperforms DP methods, showing the benefit of leveraging deviation distribution over sending a fixed portion of tensors.
Figure~\ref{fig:threshold} shows the learned threshold $\theta^r$ for our method.
The decreasing value of $\theta^r$ for ${\text{\texttt{MetaSend}}_{l}}$ indicates its focus on sending tensors with even smaller deviations during communication. As the model converges and becomes stable, the deviations decrease, resulting in ${\text{\texttt{MetaSend}}_{g}}$ generating a smaller threshold for a milder deviation distribution.

\begin{figure*}[t]
    \centering
    \setlength{\abovecaptionskip}{1mm}
    \includegraphics[width=0.85\linewidth]{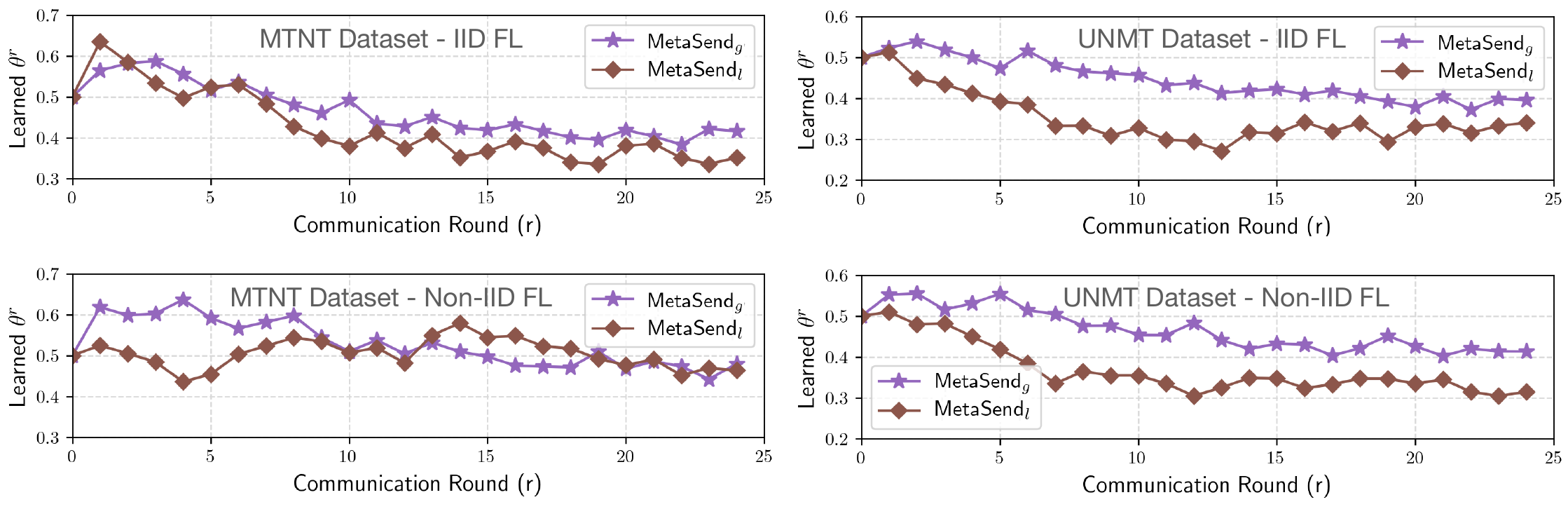}
    \caption{The learned threshold $\theta^r$ for our method across datasets and configurations shows its ability to dynamically adapt the threshold to different scenarios.}
    \label{fig:threshold}
    \vspace{-3.5mm}
\end{figure*}

\subsubsection{Parameters used in MAML module} 
\label{sec:ablation_neuron}
The default structure of our MAML module is an MLP network with one hidden layer containing 100 neurons.
To explore the impact of the number of neurons on performance, we keep the learning parameters consistent while varying the number of neurons in the hidden layer of the MAML module.
In Figure~\ref{fig:ablation_neruon_unmt}, we see that using more neurons in the MAML module generally leads to improved results in terms of sacreBLEU score and tensor savings. 
The performance gain from using more neurons is intuitive since it provides additional degrees of freedom for learning optimization.
However, it is important to note that using more neurons also incurs higher resource requirements during system construction.


\begin{figure*}[t]
    \centering
\setlength{\abovecaptionskip}{1mm}
    \includegraphics[width=0.95\linewidth]{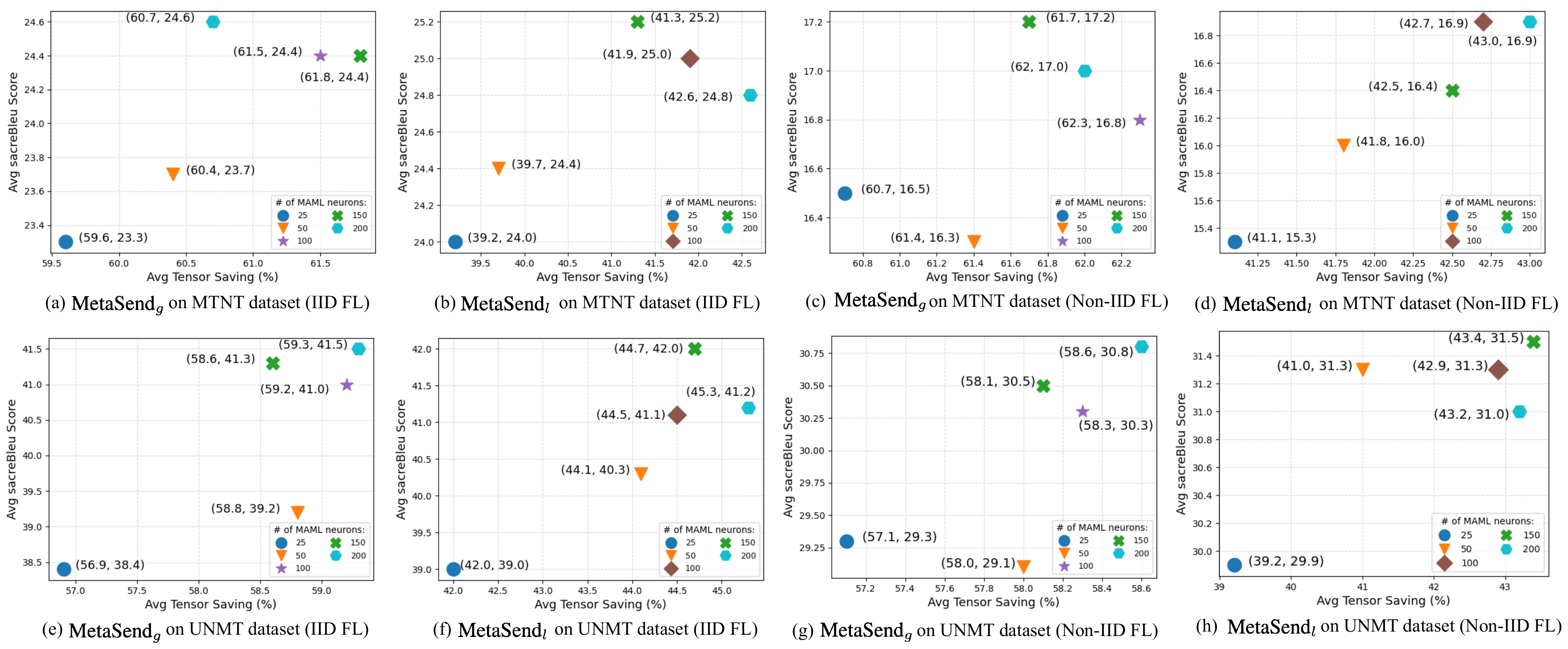}
    \caption{Average sacreBLEU scores and tensor savings for {\texttt{MetaSend}} with varying MAML module neurons. Increasing parameters improves thresholding, enhancing translation performance and efficiency.}
    \label{fig:ablation_neruon_unmt}
\end{figure*}


\subsubsection{Meta evaluation for MAML module} 
\label{sec:ablation_batch}

To examine the influence of the number of batches used to optimize our MAML module, Figure~\ref{fig:ablation_batch_unmt} shows the performance of our method with different numbers of batches used for meta-optimization. 
We see that increasing the number of samples used for MAML optimization generally results in improved translation quality and efficiency.
Naturally, using more batches will increase the exact time spent on our MAML module. 
However, the time taken by clients to send parameters for aggregation may be more critical than this optimization time, as the optimization process is performed only once in each round. 


\subsubsection{The use of pre-trained models}
\label{sec:ablation_pretrain}
Tables~\ref{tb:bleu_nopre} and ~\ref{tb:comet_nopre} show the sacreBLEU and COMET scores of each method without utilizing pre-trained knowledge for the M2M-100 model in each client. 
Specifically, each method was trained from scratch, utilizing randomly initialized weights.
Although the results of ${\text{\texttt{MetaSend}}_{l}}$ do not exhibit complete superiority over ${\text{\texttt{MetaSend}}_{g}}$ as discussed in Section~\ref{sec:result_translation}, our {\texttt{MetaSend}} methods display higher average scores overall in most cases compared with baseline approaches.
However, the overall results are relatively poor and unstable compared to the results shown in Tables~\ref{tb:bleu_score} and \ref{tb:comet_score} when pre-training is utilized.
This discrepancy can be attributed to the fact that large neural MT systems typically demand a substantial amount of data and prior knowledge to perform effectively~\cite{Chu2024RethinkingTS}.

\begin{table}[t]
\caption{Time spent for passing different numbers of batch samples to the NMT engine for meta-evaluation. It is intuitive that utilizing more samples requires additional processing time.}

\small
\begin{center}
\begin{tabular}{cc}
\hline
\textbf{Meta Evaluation}  & \textbf{Time (sec)} \\
\hline\hline

4 batches & 2.621 $\times 10^{-1}$ \\
8 batches & 5.012 $\times 10^{-1}$ \\
16 batches (default) & 8.754 $\times 10^{-1}$ \\
32 batches & 1.731  \\
64 batches &  3.682 \\

\hline
\end{tabular}

\end{center}
\vspace{-1.0pc}
\label{tb:time_ablation}
\end{table}

\begin{table*}
\caption{SacreBleu scores without pre-trained weight initialization. Bold scores highlight the best in each column. Our method outperforms baselines even with limited prior knowledge.
}

\small
\begin{center}
\begin{tabular}{ccccccccc}
\hline
\multirow{2}{*}{\textbf{Training}}& \multirow{2}{*}{\textbf{Method}}& \multicolumn{3}{c}{\textbf{MTNT}}  & \multicolumn{4}{c}{\textbf{UNMT}}  \\ 
\cmidrule(l){3-5}\cmidrule(l){6-9}
& & \textbf{En $\to$ Fr} & \textbf{En $\to$ Ja} & \textbf{Avg} & \textbf{En $\to$ Fr} & \textbf{Ar $\to$ Es} & \textbf{Ru $\to$ Zh} & \textbf{Avg}\\
\hline\hline
{Centralized}& w/o pre-trained & 18.1 & 11.0 & 14.6 & 27.9 & 25.5 & 27.2 & 26.9\\
{IID FL}& PMFL & {18.8} & 6.6 & {12.7} & 22.3 & 20.5 & {22.2}& {21.7}\\
{Non-IID FL}& PMFL  & 14.9 & 7.0 & 10.9 & {18.9} & 16.6 & {18.0} &{17.8}\\
\hline
\multirow{5}{*}{IID FL}& RandSend  & 15.5 & 5.2 & 10.4 & 20.6 & 18.3 & 17.1 & 18.7\\
& DP$_{g}$ & 16.5 & 6.3 & 11.4 & 21.3 & 20.5 & 19.0 & 20.3 \\
& DP$_{l}$ & 16.0 & 6.9 & 11.5 & 20.8 & 21.0 & 19.7 & 20.5 \\

& {MetaSend}$_g$ (ours) & \textbf{17.9}  & 7.0 & \textbf{12.5} & 20.9 & \textbf{21.7} & 18.6 & 20.4 \\
& {MetaSend}$_l$ (ours) & 17.5 & \textbf{7.5} & \textbf{12.5} & \textbf{22.7} & 20.8 & \textbf{20.4} & \textbf{21.3}\\
\hline
\multirow{5}{*}{Non-IID FL}& RandSend & 11.4 & 5.5 & 8.5 & 13.6 & 10.3 & 12.0 & 11.9 \\
& DP$_{g}$ & 14.6 & 6.6 & 10.6 & 16.3 & 14.9 & 15.0 & 15.4\\
& DP$_{l}$ & 14.9 & 6.8 & 10.9 & 16.6 & 15.4 & 15.3 & 15.8\\
& {MetaSend}$_g$ (ours) & 14.9 & 6.9 & 10.9 & \textbf{18.5} & 16.6 & \textbf{16.3} & 17.1\\
& {MetaSend}$_l$ (ours) & \textbf{15.1} & \textbf{7.3} & \textbf{11.2} & 17.7 & \textbf{18.2} & 16.0 & \textbf{17.3}\\
\hline
\end{tabular}
\vspace{-1.5pc}
\label{tb:bleu_nopre}
\end{center}
\end{table*}

\begin{table*}
\caption{COMET scores obtained with each method without using pre-trained model as initialization. The improved performance demonstrates that our method can perform robustly even in the absence of a pre-trained model.}

\small
\begin{center}
\begin{tabular}{ccccccccc}
\hline
\multirow{2}{*}{\textbf{Training}}& \multirow{2}{*}{\textbf{Method}}& \multicolumn{3}{c}{\textbf{MTNT}}  & \multicolumn{4}{c}{\textbf{UNMT}}  \\ 
\cmidrule(l){3-5}\cmidrule(l){6-9}
& & \textbf{En $\to$ Fr} & \textbf{En $\to$ Ja} & \textbf{Avg} & \textbf{En $\to$ Fr} & \textbf{Ar $\to$ Es} & \textbf{Ru $\to$ Zh} & \textbf{Avg}\\
\hline\hline
\multirow{1}{*}{Centralized}& w/o pre-trained & 0.663 & 0.670 & 0.667 & 0.733 & 0.707 & 0.746 & 0.729\\
{IID FL}& PMFL & 0.624 & 0.661 & 0.643 & {0.725} & {0.688} & 0.720 & {0.711}\\
{Non-IID FL}& PMFL & {0.626} & {0.648} & {0.637} & 0.671 & 0.633 & {0.660} & 0.655\\ 
\hline
\multirow{5}{*}{IID FL}& RandSend & 0.620 & 0.633 & 0.627 & 0.705 & 0.679 & 0.715 & 0.700\\
& DP$_{g}$ & 0.625 & 0.647 & 0.636 & 0.700 & 0.669 & 0.718 & 0.696\\	
& DP$_{l}$ & 0.631 & 0.651 & 0.641 & 0.719 & 0.673 & 0.727 & 0.706\\

& {MetaSend}$_g$ (ours) & 0.634 & \textbf{0.662} & \textbf{0.648} & 0.722 & 0.679 & \textbf{0.728} & 0.710 \\
& {MetaSend}$_l$ (ours) & \textbf{0.638} & 0.657 & \textbf{0.648} & \textbf{0.725} & \textbf{0.684} & 0.725 & \textbf{0.711}\\
\hline
\multirow{5}{*}{Non-IID FL}& RandSend & 0.602 & 0.639 & 0.621 & 0.653 & 0.630 & 0.632 & 0.638\\
& ${\text{DP}_{g}}$ & 0.611 & 0.645 & 0.628 & 0.663 & 0.624 & 0.650 & 0.646\\
& ${\text{DP}_{l}}$ & 0.613 & 0.639 & 0.626 & 0.665 & 0.637 & 0.646 & 0.649\\
& {MetaSend}$_g$ (ours) & \textbf{0.620} & 0.647 & \textbf{0.634} & \textbf{0.676} & \textbf{0.640} & 0.653 & 0.656\\
& {MetaSend}$_l$ (ours) & 0.619 & \textbf{0.648} & \textbf{0.634} & \textbf{0.676} & 0.638 & \textbf{0.659} & \textbf{0.658}\\
\hline
\end{tabular}
\vspace{-1.5pc}
\label{tb:comet_nopre}
\end{center}
\end{table*}

\subsection{Varying Data Availability}

\subsubsection{Insufficient data samples}
\label{sec:ablation_insufficient}
To mirror the limited data scenario of each client in practical FL, we performed experiments by randomly sampling a small portion of data from MTNT and UNMT datasets as training data. Specifically, we randomly selected 20\% of samples from each dataset, resulting in 1k and 10k training samples in each language direction for MTNT and UNMT datasets, respectively, while keeping the validation and test sets at the same size. All other hyperparameters, such as batch size for the NMT engine or MAML optimization, neurons in the MAML module, and learning rate, remain the same as in Section~\ref{sec:exp}.
Tables~\ref{tb:bleu_insuf} and ~\ref{tb:comet_insuf} present the translation quality of each method when trained with limited data. 
Even in scenarios with limited resources, it is evident that our {\texttt{MetaSend}} methods consistently outperform other baselines and, in some cases, achieve comparable or even slightly better performance compared to PMFL.
\begin{table*}
\caption{SacreBleu scores obtained with each method with the reduced number of training samples. We observe that our method shows improved translation quality over baselines even with limited data resources in each client.}

\small
\begin{center}
\begin{tabular}{ccccccccc}
\hline
\multirow{2}{*}{\textbf{Training}}& \multirow{2}{*}{\textbf{Method}}& \multicolumn{3}{c}{\textbf{MTNT}}  & \multicolumn{4}{c}{\textbf{UNMT}}  \\ 
\cmidrule(l){3-5}\cmidrule(l){6-9}
& & \textbf{En $\to$ Fr} & \textbf{En $\to$ Ja} & \textbf{Avg} & \textbf{En $\to$ Fr} & \textbf{Ar $\to$ Es} & \textbf{Ru $\to$ Zh} & \textbf{Avg}\\
\hline\hline
{Centralized}& w/o pre-trained & 10.7 & 6.1 & 8.4 & 13.9 & 11.8 & 12.5 & 12.7\\
{Centralized}& w/ pre-trained & 28.4 & 14.3 & 21.4 & 34.6 & 33.8 & 34.8 & 34.4\\
{IID FL}& PMFL & 27.3 & 11.5 & 19.4 & 33.6 & 33.2 & {34.4} & 33.7\\
{Non-IID FL}& PMFL & {16.6} & {8.8} & {12.7} & 19.4 & 19.6 & {17.7} & {18.9}\\ 
\hline
\multirow{5}{*}{IID FL}& RandSend & 26.1 & 10.4 & 18.3 & 31.5 & 31.3 & 32.1 & 31.6\\
& DP$_{g}$ & 27.3 & 10.9 & 19.1 & 32.6 & 32.7 & 32.3 & 32.5\\
& DP$_{l}$& 26.8 & 11.0 & 18.9 & 33.0 & 32.9 & 32.8 & 32.9\\

& {MetaSend}$_g$ (ours)& \textbf{28.0} & 11.8 & \textbf{19.9} & \textbf{34.3} & 33.0 & \textbf{34.1} & \textbf{33.8}\\
& {MetaSend}$_l$ (ours)& 27.3 & \textbf{12.0} & 19.7 & {34.2} & \textbf{33.2} & \textbf{34.1} & \textbf{33.8} \\
\hline
\multirow{5}{*}{Non-IID FL}& RandSend & 14.9 & 7.4 & 11.1 & 18.2 & 17.4 & 16.4 & 17.3\\
& DP$_{g}$ & 15.3 & \textbf{7.9} & 11.6 & 18.4 & 19.5 & 16.1 & 18.0\\
& DP$_{l}$& 15.0 & 7.3 & 11.2 & 18.9 & 19.0 & 16.4 & 18.1\\
& {MetaSend}$_g$ (ours)& 15.8 & \textbf{7.9} & \textbf{11.9} & \textbf{19.5} & 19.5 & 17.5 & 18.8\\
& {MetaSend}$_l$ (ours)& \textbf{16.0} & 7.8 & \textbf{11.9} & 19.4 & \textbf{19.7} & \textbf{17.7} & \textbf{18.9}\\
\hline
\end{tabular}
\vspace{-1.5pc}
\label{tb:bleu_insuf}
\end{center}
\end{table*}

\begin{table*}
\caption{COMET scores obtained with each method with the reduced number of training samples. We observe that our method shows improved performance over baselines even when the clients hold limited data resources.}

\small
\begin{center}
\begin{tabular}{ccccccccc}
\hline
\multirow{2}{*}{\textbf{Training}}& \multirow{2}{*}{\textbf{Method}}& \multicolumn{3}{c}{\textbf{MTNT}}  & \multicolumn{4}{c}{\textbf{UNMT}}  \\ 
\cmidrule(l){3-5}\cmidrule(l){6-9}
& & \textbf{En $\to$ Fr} & \textbf{En $\to$ Ja} & \textbf{Avg} & \textbf{En $\to$ Fr} & \textbf{Ar $\to$ Es} & \textbf{Ru $\to$ Zh} & \textbf{Avg}\\
\hline\hline
{Centralized}& w/o pre-trained & 0.355 & 0.499 & 0.427 & 0.398 & 0.413 & 0.379 & 0.397\\
{Centralized}& w/ pre-trained & 0.698 & 0.687 & 0.693 & 0.764 & 0.745 & 0.732 & 0.747 \\
{IID FL}& PMFL  & 0.687 & 0.683 & 0.685 & 0.750 & {0.736} & 0.745 & {0.744}\\
{Non-IID FL}& PMFL  & {0.640} & 0.661 & 0.651 & 0.677 & 0.664 & 0.671 & 0.671\\ 
\hline
\multirow{5}{*}{IID FL}& RandSend & 0.666 & 0.655 & 0.661 & 0.733 & 0.719 & 0.727 & 0.726 \\
& DP$_{g}$& 0.701 & 0.678 & 0.690 & 0.746 & 0.727 & 0.739 & 0.737\\
& DP$_{l}$& 0.691 & 0.680 & 0.686 & 0.741 & 0.729 & 0.733 & 0.734\\
& {MetaSend}$_g$ (ours)& \textbf{0.711} & 0.680 & \textbf{0.696} & \textbf{0.752} & \textbf{0.736} & 0.743 & \textbf{0.744}\\
& {MetaSend}$_l$ (ours)& 0.687 & \textbf{0.684} & 0.685 & 0.750 & \textbf{0.736} & \textbf{0.747} & 0.743\\
\hline
\multirow{5}{*}{Non-IID FL}& RandSend & 0.618 & 0.640 & 0.629 & 0.649 & 0.624 & 0.634 & 0.636\\\
& DP$_{g}$ & 0.633 & 0.654 & 0.644 & 0.673 & 0.649 & 0.659 & 0.660\\
& DP$_{l}$ & 0.631 & 0.659 & 0.645 & 0.680 & 0.653 & 0.678 & 0.670\\
& {MetaSend}$_g$ (ours)& \textbf{0.640} & \textbf{0.670} & \textbf{0.655} & 0.659 & 0.670 & \textbf{0.711} & 0.680 \\
& {MetaSend}$_l$ (ours)& 0.638 & \textbf{0.670} & 0.654 &\textbf{0.685} & \textbf{0.688} & 0.685 & \textbf{0.686}\\
					
\hline
\end{tabular}
\vspace{-1.5pc}
\label{tb:comet_insuf}
\end{center}
\end{table*}

{
\subsubsection{Low-resource language scenarios}
In addition to scenarios with sparse data samples for every client, we also consider the case of data imbalance, where only one of the clients is designated as low-resource. 
Specifically, we randomly select 20\% of the samples from the En $\to$ Ja language pair in the MTNT dataset and 20\% from the Ru $\to$ Zh language pair in the UNMT dataset, while maintaining the full training samples for other clients and keeping the validation and test sets unchanged. 
This selection creates a scenario where a specific language pair (En $\to$ Ja and Ru $\to$ Zh) becomes low-resource. Table~\ref{tb:rebutal_lowresource} presents the results under this setting. It is evident that the NMT model achieves limited performance for the low-resource client, and the overall performance is also negatively impacted due to this client's constraints, as compared to the full results for PMFL shown in Table~\ref{tb:bleu_score}. However, in comparison to the baseline methods in Table~\ref{tb:rebutal_lowresource}, our adaptive thresholding approach delivers the best performance. This underscores the effectiveness of our adaptive thresholding approach in mitigating the challenges posed by low-resource clients, resulting in better overall performance.
}

\begin{table*}
\caption{SacreBLEU scores obtained for each method under the low-resource scenario, using 20\% of the En $\to$ Ja data samples for MTNT and 20\% of the Ru $\to$ Zh data samples for UNMT, respectively.}

\small
\begin{center}
\begin{tabular}{cccccccc}
\hline
\multirow{2}{*}{\textbf{Method}} &\multicolumn{3}{c}{\textbf{MTNT}}  &\multicolumn{4}{c}{\textbf{UNMT}}  \\ 
\cmidrule(l){2-4}\cmidrule(l){5-8}

 & \textbf{En $\to$ Fr} & \textbf{En $\to$ Ja} &  \textbf{Avg} & \textbf{En $\to$ Fr} & \textbf{Ar $\to$ Es} & \textbf{Ru $\to$ Zh} & \textbf{Avg}\\
\hline\hline

 PMFL & 23.1 & 15.6 & 19.4 & 27.9 & 28.4 & 22.9 & 26.4   \\
\hline
 RandSend &14.9 & 9.9 & 12.4 &21.9 & 20.6 & 15.7 & 19.4   \\
 DP$_{g}$ &17.3 & 11.6 & 14.5 & 25.4 & 23.9 & 17.3 & 22.2\\
 DP$_{l}$ &18.1 & 12.5 & 15.3 &25.8 & 24.2 & 17.8 & 22.6\\
 {\tt{MetaSend}$_g$} (ours) &19.7 & \textbf{13.9} & 16.8 & 25.8 & 23.9 & 17.5 & 22.4\\
 {\tt{MetaSend}$_l$} (ours) & \textbf{20.1} & 13.6 & \textbf{16.9}& \textbf{26.1} & \textbf{24.7} & \textbf{18.0} & \textbf{22.9}\\
\hline
\end{tabular}
\vspace{-1.5pc}
\label{tb:rebutal_lowresource}
\end{center}
\end{table*}

\subsection{Compatibility with other Pruning Methods}
We have shown {\texttt{MetaSend}}'s ability to learn adaptive thresholds for identifying relevant parameters for communication across various scenarios. To further explore its potential, we examined its integration with general model pruning methods~\cite{Tyagi2019SecondOT,Yeom2019PruningBE,Vadera2020MethodsFP}. Given the substantial number of parameters and complex model blocks in language models, directly implementing the additional computations proposed in~\cite{Tyagi2019SecondOT,Yeom2019PruningBE} could significantly increase client computational overhead. Instead, we adapt their concepts by introducing an auxiliary multi-layer perceptron (MLP) layer at the output of the language model to identify redundant parameters, providing further refinement to the parameters pruned by {\texttt{MetaSend}}. Similar to the importance concept introduced in \cite{Yeom2019PruningBE}, the MLP takes the magnitudes of the remaining tensors as input and determines whether they should be further pruned.
Acting as a second filter, this MLP predicts which parameters are redundant and can potentially be removed to enhance communication efficiency.

Table~\ref{tb:rebuttal_mlpprune} presents the results of integrating our method with the MLP-based pruning approach on the MTNT and UNMT datasets. With this second filtering step, the MLP-based pruning method further reduces parameter transmissions, although the impact on translation performance varies. For {\texttt{MetaSend}}$_{g}$, it has no noticeable impact on translation quality, whereas for {\texttt{MetaSend}}$_{l}$, it achieves greater parameter savings at the cost of a slight reduction in translation performance. Overall, these results demonstrate the compatibility of our method with other pruning techniques to further optimize communication efficiency, while also highlighting that {\texttt{MetaSend}} itself effectively identifies and prioritizes the most relevant parameters.

\begin{table*}
\caption{Improvements obtained from integrating our method with MLP-based pruning.}

\small
\begin{center}

\begin{tabular}{cccccc}
\hline
\multirow{2}{*}{\textbf{Training}}& \multirow{2}{*}{\textbf{Method}}  & \multicolumn{2}{c}{\textbf{MTNT}} & \multicolumn{2}{c}{\textbf{UNMT}}  \\ 
\cmidrule(l){3-4}\cmidrule(l){5-6}

& & \textbf{Translation Performance} & \textbf{Tensor Saving} & \textbf{Translation Performance} & \textbf{Tensor Saving}\\
\hline

\multirow{5}{*}{IID FL}& {MetaSend}$_g$ & 24.4 & 61.5\% & 41.0 & 59.2\% \\
& {MetaSend}$_g$ + MLP & 24.3 & +5.4\% & 41.1 & +7.9\%  \\
\cmidrule(l){2-6}
& {MetaSend}$_l$ & 25.0 & 41.9\% & 41.1 & 44.5\% \\
& {MetaSend}$_l$ + MLP & 24.7 & +10.0\% & 40.8 & +9.7\% \\
\hline
\multirow{5}{*}{Non-IID FL}& {MetaSend}$_g$ & 16.8 & 62.3\% & 30.3 & {58.3\%}\\
& {MetaSend}$_g$ + MLP & 16.8 & +6.9\% & {30.5} & {+8.2\%} \\
\cmidrule(l){2-6}
& {MetaSend}$_l$ & 16.9  & 42.7\% &{31.3} & {43.9\%} \\
& {MetaSend}$_l$ + MLP & 16.4  & +9.2\% & {30.9} & {+11.3\%} \\
			
\hline
\end{tabular}
\label{tb:rebuttal_mlpprune}
\end{center}
\end{table*}



\begin{figure*}[t]
    \centering
\setlength{\abovecaptionskip}{1mm}
    \includegraphics[width=0.95\linewidth]{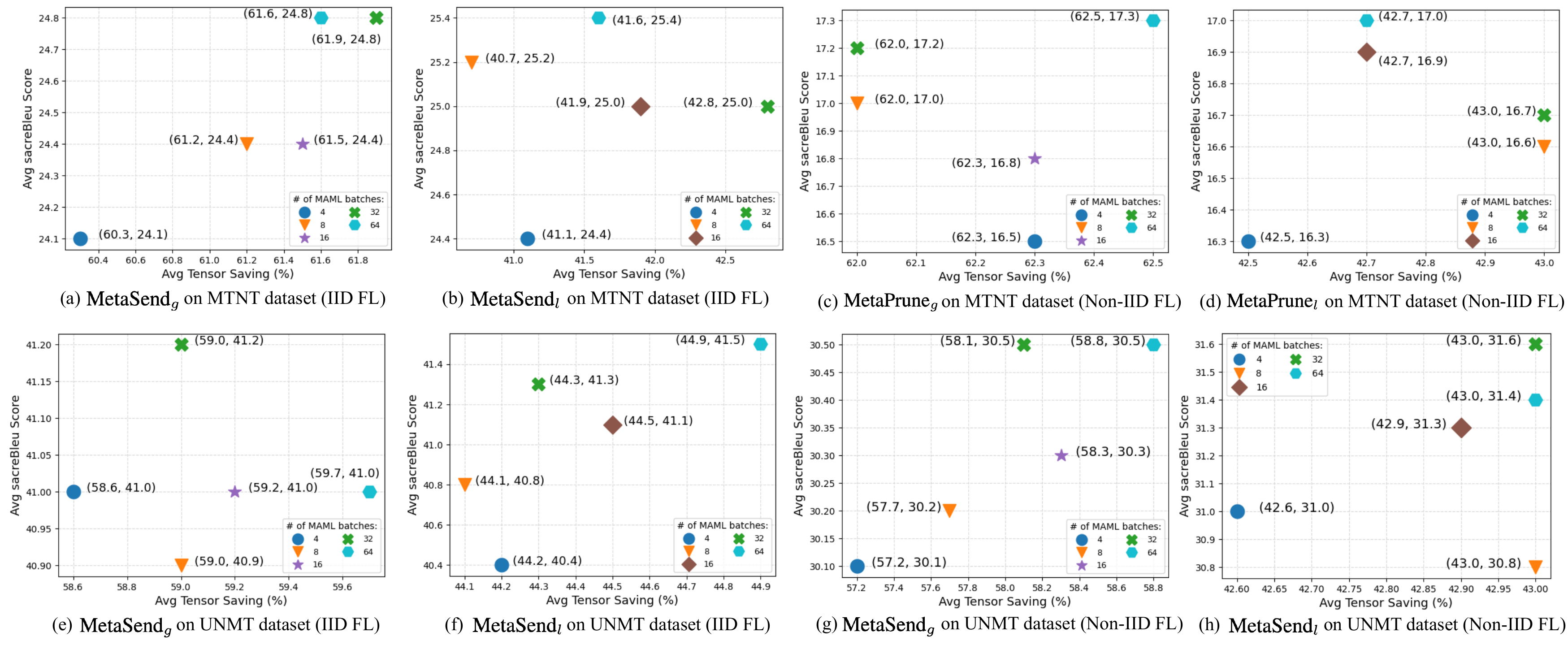}
    \caption{Average sacreBLEU scores and tensor savings for {\texttt{MetaSend}} with varying MAML batch sizes. More data batches for meta-updates generally enhance the MAML module's inference, improving translation performance and efficiency.}
    \label{fig:ablation_batch_unmt}
    \vspace{-3mm}
\end{figure*}

\section{Conclusion}
We conducted the first analysis on the efficiency of federated multilingual NMT.
To address the practical challenges that arise in this setup, we proposed {\texttt{MetaSend}}, which selects tensors for transmission that are most critical to the NMT. 
Our observation shows that designing a dynamic sending threshold to control the transmitted parameters in each client is crucial during FL communication.
By adaptively learning the sending threshold in each FL round based on meta-learning, we saw that our designed {\texttt{MetaSend}} not only improves communication efficiency, but also effectively captures the NMT threshold for sending.
Extensive experiments on two datasets showed that {\texttt{MetaSend}} outperforms existing baselines regarding machine translation quality and significantly reduces communication costs in FL, confirming its advantage in practical federated NMT settings.




\bibliographystyle{IEEEtran}
\bibliography{custom}

\appendices
\clearpage

\vfill

\end{document}